\documentclass{article}

\usepackage{arxiv}

\usepackage[utf8]{inputenc} 
\usepackage[T1]{fontenc}    
\usepackage{hyperref}       
\usepackage{url}            
\usepackage{booktabs}       
\usepackage{amsfonts}       
\usepackage{nicefrac}       
\usepackage{microtype}      
\usepackage{lipsum}		
\usepackage{graphicx}
\usepackage{doi}

\usepackage{lineno,hyperref}
\modulolinenumbers[5]
\usepackage{multirow}
\usepackage{mathtools}
\usepackage{amsthm}
\usepackage{xfrac}
\usepackage{nicefrac}
\usepackage{todonotes}
\usepackage{amsthm,amsmath}
\usepackage[utf8]{inputenc}
\usepackage{subcaption}
\usepackage{algpseudocode}
\usepackage{algorithm}
\usepackage{url}

\usepackage{tikz}
\usetikzlibrary{arrows,positioning}

\def\BibTeX{{\rm B\kern-.05em{\sc i\kern-.025em b}\kern-.08em
    T\kern-.1667em\lower.7ex\hbox{E}\kern-.125emX}}
\usepackage{balance}
\begin{document}
\title{Balancing Performance and Energy Consumption of Bagging Ensembles for the Classification of Data Streams in Edge Computing}


\author{ \href{https://orcid.org/0000-0003-4029-2047}{\includegraphics[scale=0.06]{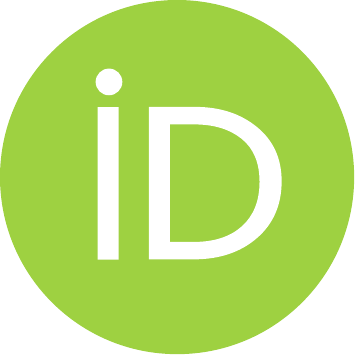}\hspace{1mm}Guilherme Weigert Cassales}\thanks{Use footnote for providing further
		information about author (webpage, alternative
		address)---\emph{not} for acknowledging funding agencies.} \\
	University of Waikato \\
	Hamilton, New Zealand \\
	\texttt{gwcassales@gmail.com} \\
	\And
	\href{https://orcid.org/0000-0002-5276-637X}{\includegraphics[scale=0.06]{orcid.pdf}\hspace{1mm}Heitor Murilo Gomes} \\
	University of Waikato \\
	Hamilton, New Zealand \\
	\texttt{heitor.gomes@waikato.ac.nz} \\
	\And
	\href{https://orcid.org/0000-0002-8339-7773}{\includegraphics[scale=0.06]{orcid.pdf}\hspace{1mm}Albert Bifet} \\
	University of Waikato \\
	Hamilton, New Zealand \\
	\texttt{abifet@waikato.ac.nz} \\
	\And
	\href{https://orcid.org/0000-0002-3732-5787}{\includegraphics[scale=0.06]{orcid.pdf}\hspace{1mm}Bernhard Pfahringer} \\
	University of Waikato \\
	Hamilton, New Zealand \\
	\texttt{bernhard@waikato.ac.nz} \\
	\And
	\href{https://orcid.org/0000-0003-1273-9809}{\includegraphics[scale=0.06]{orcid.pdf}\hspace{1mm}Hermes Senger} \\
	Department of Computing\\
	Federal University of São Carlos\\
	São Carlos, Brazil \\
	\texttt{hermes@ufscar.br} \\
}

\renewcommand{\shorttitle}{\textit{arXiv} Template}

\hypersetup{
pdftitle={Balancing Performance and Energy Consumption of Bagging Ensembles for the Classification of Data Streams in Edge Computing},
pdfsubject={q-bio.NC, q-bio.QM},
pdfauthor={Guilherme Cassales, Heitor Gomes, Albert Bifet, Bernhard Pfahringer, Hermes Senger},
pdfkeywords={Edge computing, Machine learning, Data stream classification, Ensembles, Energy consumption},
}


\maketitle

\begin{abstract}
In recent years, the Edge Computing (EC) paradigm has emerged as an enabling factor for developing technologies like the Internet of Things (IoT) and 5G networks, bridging the gap between Cloud Computing services and end users, supporting low latency, mobility, and location awareness to delay-sensitive applications. Most solutions in EC employ machine learning (ML) methods to perform data classification and other information processing tasks on continuous and evolving data streams.
Usually, such solutions have to cope with vast amounts of data that come as data streams while balancing energy consumption, latency, and the predictive performance of the algorithms. 
Ensemble methods achieve remarkable predictive performance when applied to evolving data streams due to the combination of several models and the possibility of selective resets. This work investigates strategies for optimizing the performance (i.e., delay, throughput) and energy consumption of bagging ensembles to classify data streams. The experimental evaluation involved six state-of-art ensemble algorithms (OzaBag, OzaBag Adaptive Size Hoeffding Tree, Online Bagging ADWIN, Leveraging Bagging, Adaptive RandomForest, and Streaming Random Patches) applying five widely used machine learning benchmark datasets with varied characteristics on three computer platforms. Such strategies can significantly reduce energy consumption in 96\% of the experimental scenarios evaluated. 
Despite the trade-offs, it is possible to balance them to avoid significant loss in predictive performance.
\end{abstract}


\section{Introduction}


{
Edge Computing (EC) is an enabling paradigm for developing technologies like the Internet of Things (IoT), 5G, online gaming, augmented reality (AR), vehicle-to-vehicle communications, smart grids, and real-time video analytics. In essence, EC brings the services and utilities of cloud computing closer to the end user, providing low latency, location awareness, and better efficiency for delay-sensitive mobile applications \cite{khan2019edge}.
To accomplish this, computational resources are placed at the possible closest location to the mobile devices, with moderate servers’ capabilities placed at the edge of the network to achieve necessary user centric requirements \cite{Shakarami2020}.


Many applications like smart grids with a large number of sensors continuously collect data from the surrounding environment across medium to large geographical areas \cite{Yu2011}.
Because most sensors are resource-constrained, they cannot run more expensive or data-intensive tasks like machine learning (ML) algorithms. In this case, it is more efficient to send these data to more powerful resources nearby (placed on the edge of the network) for faster processing (i.e., with low latency) \cite{Lopes20, Coutinho2020}. 
Another example is wearable technologies,   
which can depend on energy- and resource-constrained mobile devices or low-end servers placed in the nearby to process their data streams  \cite{Jin2021}.

}



In these scenarios, a remarkable trend is the increasing adoption of ML techniques to execute tasks like data classification, spam filtering, anomaly detection in network traffic for cybersecurity, real-time image classification and segmentation, driving support applications, autonomous driving vehicles, and many others. Among ML techniques, ensembles of classifiers have demonstrated remarkable predictive performance for the classification of data streams~\cite{gomes2017survey}.

Historically, ML research focused on improving predictive performance without constraints for computational resources and energy consumption. However, the need to cope with these dynamic environments with potentially infinite data streams with non-stationary behavior, and to run ML algorithms either on mobile devices or low-end servers in the edge create additional challenges. 
On the algorithmic side, the ML field is shifting towards data stream learning to face such challenges where requirements like single pass, response time, and constant memory usage are imposed~\cite{gama2014survey}. Nevertheless, little effort has been made to reduce the energy consumption of such algorithms \cite{energy-albert}.

To address this gap, we have elaborated on previous research~\cite{HPCC,IS} to investigate how to optimize the classification of data streams with bagging ensembles with respect to energy efficiency, time performance (i.e., delay, and throughput), and predictive performance. The main contributions of this paper can be summarized as follows: 
\begin{enumerate}
    \item Present the premier study that applies {\it mini-batching} (proposed in \cite{HPCC,IS}) for improving the energy efficiency of bagging ensembles;
    \item Identify the trade-offs between timer performance, energy efficiency, and predictive performance of bagging ensembles for the classification of online data streams;
    \item Demonstrate how to achieve a good balance of these trade-offs to obtain significant gains on performance and energy efficiency, at cost of negligible loss in predictive performance on a realistic stream processing testbed with ({\it i}) six state of the art ensemble algorithms, ({\it ii}) five widely used dataset benchmarks,  ({\it iii}) three computer architectures ranging from micro- to middle-end servers, and ({\it iv}) three levels of stream workloads intensities (i.e., 10\%, 50\%, and 90\% of the maximum server capacity);
    \item Our results present significant gains in 96\% of the evaluated cases.
\end{enumerate}

The remainder of this article is organized as follows, \autoref{sec:relatedwork} discusses the related works. The state-of-art bagging ensemble algorithms are described in \autoref{sec:ensembles}. We present our proposal for applying mini-batching for parallel implementations of bagging ensemble algorithms in \autoref{sec:proposal}, followed by the experimental evaluation in \autoref{sec:experiments}. Finally, our conclusion is presented in \autoref{sec:conclusions}.
\section{Related work}
\label{sec:relatedwork}

{
Computing nodes, from smallest devices to high-end servers comprise several components, most of which can be optimized to save energy \cite{Orgerie2014}. 
From atomic block components like a simple functional unit within a chip to CPU cores, disks, network interface cards (NICs) and entire boards can be put in sleep or idle states for saving power when not delivering services
to achieve proportional computing \cite{4404806}. Dynamic power management (DPM) encompasses a set of techniques that achieve energy-efficient by selectively turning off (or reducing the performance of) system components when they are idle (or partially unexploited) \cite{benini2000survey}. Dynamic Voltage and Frequency Scaling (DVFS) define several levels of frequency at which a processor can operate, where lower frequencies become slower to save power \cite{Orgerie2014, snowdon2005power}.   Sensors and wearable devices that use small-sized
batteries can also be optimized to save power while delivering functionalities including sensing, storage, and computation \cite{Jin2021}.
The work in \cite{Coutinho2020} combines the use of DVFS and DPM for optimizing performance and energy savings of single-board computers using Pareto frontier for multi-criteria optimization of several computing intensive kernels. {
Despite DVFS can reduce energy consumption, scaling the CPU clock frequency also affects the performance of other applications running in the same node.}} 

A modular, scalable, and efficient FPGA-based implementation of kNN for System on Chip devices is presented in \cite{Vieira-KNN}. The solution shows improvements of 60X in execution time and 50X in energy efficiency due to the low power consumption of FPGAs. Despite of its outstanding performance, the contribution is specific in terms of algorithm and hardware.   

The work in \cite{Martin15} emphasizes energy consumption and energy efficiency as important factors to consider during data mining algorithm analysis and evaluation. The work extended the CRISP (Cross Industry Standard Process for Data Mining) framework to include energy consumption analysis, demonstrating how energy consumption and accuracy are affected when varying the parameters of the Very Fast Decision Tree (VFDT) algorithm. The results indicate that energy consumption can be reduced by up to 92.5\% while maintaining accuracy.

In \cite{Amezzane19}, the authors analyze power consumption for both batch and online data stream learning. They experimented with three online and three batch algorithms. Among their conclusions is the finding that the CPU consumes up to 87\% of the total energy. 
Although evaluating online learners, this work tested only single model classifiers. 

{
Aiming to reduce the memory cost, the work in  \cite{Costa2018} proposed the Strict VFDT (SVFDT), an algorithm which extends the VFDT. Designed for memory constrained devices, 
the algorithm minimizes unnecessary tree growth, substantially reducing memory usage and execution time, while keeping competitive predictive performance. A comparison of the four-way relationship among time efficiency, energy consumption, predictive performance, and memory costs is presented in \cite{Lopes20}. The comparison is made by tuning the hyper-parameters of VFDT, SVFDT, and SVFDT with OLBoost. The work demonstrated that the most complex method delivers the best predictive performance at the expense of worse memory and energy performance. 

In \cite{energy-albert}, the authors presented an energy-efficient approach to real-time prediction with high levels of accuracy called \textit{nmin adaptation}. This reduces the energy-consumption of Hoeffding Trees ensembles by adapting the number of instances required to create a split. This method can reduce energy consumption by 21\% on average with a small impact on accuracy. They also presented detailed theoretical energy models for ensembles of Hoeffding trees and a generic approach to creating energy models applicable to any class of algorithms. Although these works propose more energy efficient algorithms, the benefit achieved in these works are limited to the specific algorithms. In contrast, our work focus on optimizations which can be applied to several bagging ensembles. }

{
In summary, the related works are still scarce, and mainly focus either in designing power efficient algorithms or optimizing existing ones. Our purpose is to study the benefits and trade-offs between time performance (e.g., latency and throughput), energy efficiency and predictive performance of applying mini-batching, an optimization technique (proposed in \cite{CASSALES2021260}) which can be applied to bagging ensembles. 
Mini-batching is orthogonal to any {\it ad hoc} optimization of specific learning algorithms within the ensemble like the proposals found in \cite{Costa2018,energy-albert,Martin15,Vieira-KNN}. Being orthogonal, mini-batching can be combined with such {\it ad hoc} optimizations that focus on a specific learner algorithms within the ensemble. This combination, however, is out of this work’s scope. }




\section{Bagging ensembles for stream processing }
\label{sec:ensembles}





In many applications, learning algorithms have to cope with dynamic environments that collect potentially unlimited data streams.
Formally, a data stream $S$ is a massive sequence of data elements $x_1, x_2, \dots, x_n$ that is, $S={\{x_i\}}_{i=1}^n$, which is potentially unbounded (n → $\infty$) \cite{silva2013data}.
As mentioned before, stream processing algorithms have additional requirements, which may be related to memory, response time, or a transient behavior presented by the data stream. In this context, one of the most widely used algorithms is the Hoeffding Tree~\cite{HOEFFDING_TREE}.

The Hoeffding Tree (HT) is an incremental tree designed to cope with massive data streams. Thus, it can create splits with reasonable confidence in the data distribution while having very few instances available. This is possible because of the Hoeffding Bound (HB), which states that with probability $1 - \delta$, the true mean of the variable is at least within $\pm\epsilon$ of the average of the observed variable. 

\begin{equation}
    \epsilon = \sqrt{  \frac{R^2\ ln(1/\delta)}{2n}  },
\end{equation}  
where $r$ is a real-valued random variable with a range $R = r_{max} - r_{min}$ (i.e., the subtraction of the maximum and minimum values of $r$) considering the $n$ independent observations of $r$.

\begin{figure*}[ht]
\centering
\begin{tikzpicture}[]
\draw (-2.65,2.75) rectangle (-1.0,1.75) node[pos=0.5] (l1) {...}; 
\draw (-2.65,1.75) rectangle (-1.0,0.75) node[pos=0.5] {$x^{t-1},y^{t-1}$};
\draw (-2.65,0.75) rectangle (-1.0,-0.25) node[pos=0.5] (data) {$x^t,y^t$};
\draw (-2.65,-0.25) rectangle (-1.0,-1.25) node[pos=0.5] {$x^{t+1},y^{t+1}$};
\draw (-2.65,-1.25) rectangle (-1.0,-2.25) node[pos=0.5] {...}; 
\node[rotate=90] at (-3.05,0.1) {\large  Stream};
\draw (-0.75,3.5) rectangle (10,-3.0);
\draw (-0.5,1) rectangle (1.5,-0.5) node[pos=0.5,align=center] (input)  {Poisson\\distribution\\weighting};

\draw (3,3) rectangle (5,2) node[pos=0.5] (tree1) {learner \#1};
\draw (3,1.5) rectangle (5,0.5) node[pos=0.5] (tree2) {learner \#2};
% ...
\draw[white] (3,0) rectangle (5,-1) node[black, pos=0.5] {...};
\draw (3,-1.5) rectangle (5,-2.5) node[pos=0.5] (treen) {learner \#n};

\draw (7,0.75) rectangle (9.5,-0.75) node[pos=0.5, align=center] (vote) {Majority vote\\ aggegation};
\draw (11.5,0) ellipse (1cm and 0.5cm) node[anchor=center] {Decision};
\draw[-latex] (data)  -- (input);
\draw[-latex] (1.5,0.25)  -- (3,2.5);
\draw[-latex] (1.5,0.25)  -- (3,1);
\draw[-latex] (1.5,0.25)  -- (3,-2);
\draw[-latex] (5,2.5) -- (7,0);
\draw [-latex] (5,1) -- (7,0);
\draw[-latex] (5,-2) -- (7,0);
\draw[-latex] (9.5,0) -- (10.4,0);
\end{tikzpicture}
\caption{Example of a Bagging Ensemble organization.}
\label{fig:ensemble}
\end{figure*}
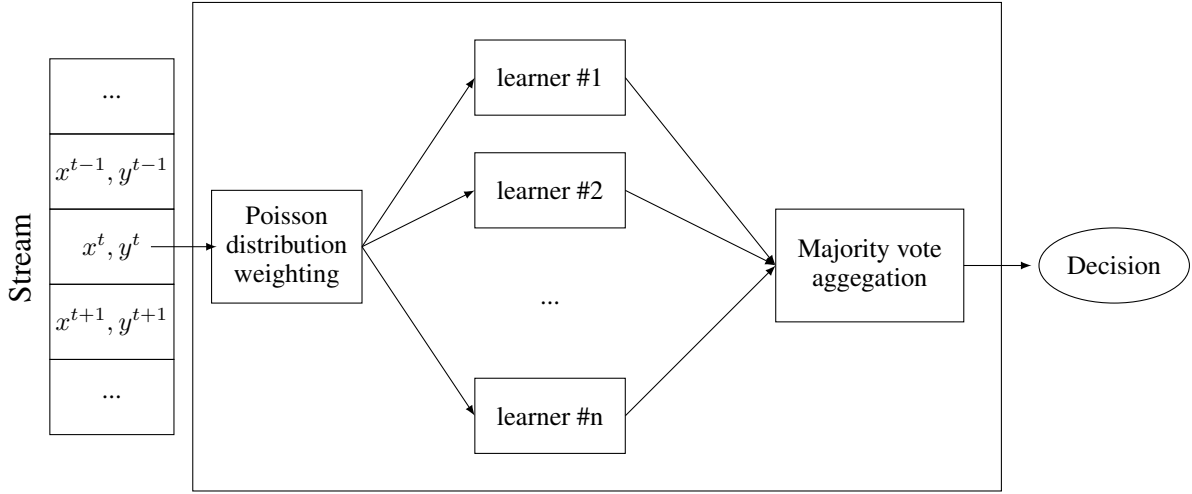

One of the shortcomings of the HT is that a single tree is considered a `weak' model in the sense that a single tree cannot accurately model complex learning problems. One approach to circumvent such `weakness' is to ensemble several models. A popular strategy to create an ensemble of learners is Bagging~\cite{Breiman1996}. 
Although Breiman proposed Bagging and its variants (e.g., Random Forest) more than 20 years ago \cite{Breiman1996}, they are still used to this day as an effective method to reduce error without resorting to intricate models, such as deep neural networks, that are not trivial to train and fine-tune.
In contrast to Boosting~\cite{ADABOOST}, Bagging does not create dependency among the base models, facilitating the parallelization of the method in an online fashion. 
Fig. \ref{fig:ensemble} {
depicts the streaming adaptation of Bagging that calculates different weights using a Poisson distribution to simulate the batch bootstrapping process and preserves independence among the learners.
The Poisson distribution effectively creates several different subsets of the data that allow repeatedly training with some instances while ignoring others in the training phase.
} 
Besides that, Bagging variants yield higher predictive performance in the streaming setting than Boosting or other ensemble methods that impose dependencies among its base models. This phenomenon is present in several empirical experiments \cite{OzaBag,ozabagadwin,Levbag,Gomes2017}. This behavior can be attributed to the difficulty of effectively modeling the dependencies in a streaming scenario, as noted in \cite{gomes2017survey}.

Next, we present a summary description of six ensemble algorithms that evolved from the original Bagging to an online (streaming) setting by Oza and Russell~\cite{OzaBag}.
Despite other decision tree algorithms~\cite{EFDT}, the HT algorithm is often chosen as the base model for the online bagging algorithms. The HT is often a common denominator, it may not be the most accurate individually, but it does yield reasonable predictive performance without requiring too many computational resources. Our experiments employ the HT as the base model for all six algorithms used to evaluate our mini-batching strategy.

\textbf{Online Bagging (OzaBag - OB)}~\cite{OzaBag} is an incremental adaptation of the original Bagging algorithm. The authors demonstrate how the process of bootstrapping can be adapted to an online setting using a Poisson($\lambda=1$) distribution. In essence, instead of sampling with replacement from the original training set, in Online Bagging, the Poisson($\lambda=1$) is used to assign weights to each incoming instance. These weights represent the number of times an instance will `repeated' to simulate bootstrapping. One concern with using $\lambda=1$ is that about $37\%$ of the instances will receive weight 0, thus not being used to train, which is desired to approximate it to the offline version of Bagging, but may be detrimental to an online learning setting~\cite{gomes2017survey}. Therefore, other works~\cite{Levbag,Gomes2017} increase the number of times an instance is used for training by increasing the $\lambda$ parameter.

\textbf{OzaBag Adaptive Size Hoeffding Tree (OBagASHT)} \cite{ozabagadwin} combines the OzaBag with Adaptive-Size Hoeﬀding Trees (ASHT). The new trees have a maximum number of split nodes and some policies to prevent the tree from growing bigger than this parameter (i.e. deleting some nodes). This algorithms' objective was to improve predictive performance by enforcing the creation of different trees. Effectively, diversity is created by having different reset-speed trees in the ensemble, according to the maximum size. The intuition is that smaller trees can adapt more quickly to changes, and larger trees can provide better performance on data with little to no changes in distribution. Unfortunately, in practice, this algorithm did not outperform variants that relied on other mechanisms for adapting to changes, such as resetting learners periodically or reactively~\cite{gomes2017survey}.

\textbf{Online Bagging ADWIN (OBADWIN)}~\cite{ozabagadwin} combines OzaBag with the ADAptive WINdow (ADWIN)~\cite{ADWIN} change detection algorithm. 
When a change is detected, the classifier with the lowest predictive performance is replaced by a new classifier. 
ADWIN keeps a variable-length window of recently seen items. The property that the window has the maximal length statistically consistent with the hypothesis that there has been no change in the average value inside the window. This implies that the average over the existing window can be reliably taken as an estimation of the current average in the stream at any time, except for a very small or very recent change that is still not statistically visible.

\textbf{Leveraging Bagging (LBag)}~\cite{Levbag} extends OBADWIN by increasing the $\lambda$ parameter of the Poisson distribution to $6$, effectively causing each instance to have a higher weight and be used for training more often. 
In contrast to OBADWIN, LBag maintains one ADWIN detector per model in the ensemble and independently resets the models. 
This approach leverages the predictive performance of OBADWIN by merely training each model more often (higher weight) and resetting them individually. One drawback of LBag compared with OB and OBADWIN is that it requires more memory and processing time since the base models are trained more often, and there are more instances of ADWIN. 
In \cite{Levbag}, the authors also attempted to further increase the diversity of LBag by randomizing the output of the ensemble via random output codes. However, this approach was not very successful compared to maintaining a deterministic combination of the models' outputs. 

\textbf{Adaptive Random Forest (ARF)} is an adaptation of the original Random Forest algorithm~\cite{RANDOM_FORESTS} to the streaming setting. Random Forest can be seen as an extension of Bagging, where further diversity among the base models (decision trees) is obtained by randomly choosing a subset of features to be used for further splitting leaf nodes. 
ARF uses the incremental decision tree algorithm Hoeffding tree~\cite{HOEFFDING_TREE} and simulates resampling as in LBag, i.e., Poisson($\lambda=6$). The Adaptive part of ARF stems from the change detection and recovery strategies based on detecting warnings and drifts per tree in the ensemble. After a warning is signaled, another model is created (namely, a `background tree') and trained without affecting the ensemble predictions. If the warning escalates to a drift signal, then the associated tree is replaced by its background tree. Notice that in the worst case, the number of tree models in ARF can be at most double the total number of trees due to the background trees. However, as noted in \cite{Gomes2017} the co-existence of a tree and its background tree is often short-lived. 

\textbf{Streaming Random Patches (SRP)} \cite{SRP} is an ensemble method specially adapted to stream classification, which combines random subspaces and online bagging. SRP is not constrained to a specific base learner as ARF since its diversity inducing mechanisms are not built-in the base learner, i.e., SRP uses global randomization while ARF uses local randomization. 
Even though, in \cite{SRP} all the experiments focused on Hoeffding trees and showed that SRP could produce deeper trees, which may lead to increased diversity in the ensemble.

\section{Mini-batching for improving the performance of ensembles}
\label{sec:proposal}


Although all the learners who compose an ensemble may be homogeneous in type, each has its own (and different) model. For instance, all learners implement a Hoeffding Tree that grows in a different shape and can change over time. 
One advantage of such methods is that task parallelism can naturally be applied as the underlying classifiers in bagging ensembles execute independently from each other and without communication.

Algorithm \ref{alg:highlevel-new} depicts a task-parallel-based implementation.
This version improves the performance of the current parallel implementation of the ARF algorithm~\cite{Gomes2017}, in the latest version in MOA~\cite{bifet2010moa}, by reusing the data structures and avoiding the costs of allocating new ones for every instance to be processed.

\begin{algorithm}
  \caption{High level parallel algorithm}
  \label{alg:highlevel-new}
  \begin{algorithmic}[1]
    \State {\bf Input}: an ensemble $E$, $num\_threads$, a data stream $S$
    \State $P \gets Create\_service\_thread\_pool(num\_threads)$
    \State $T \gets Create\_trainers\_collection(E)$
    \For {each arriving instance $I$ in stream $S$}
    \State $E$.classify($I$)
    \For  {each trainer $T_i$ in trainers $T$} 
    \State $k \gets poisson(\lambda)$
    \State $T_i.update(I, k)$
    \EndFor
    \For {all trainers $T$} {\bf in parallel}
    \State $W\_inst \gets I * k$
    \State $Train\_on\_instance(W\_inst)$
    \EndFor
    \If {change detected}
    \State $reset\_classifier$
    \EndIf
    \If {$ElapsedTime > Timeout$}
    \State {\textbf{break}}
    \EndIf
    \EndFor
  \end{algorithmic}
\end{algorithm}
In lines 2-3, we start a thread pool and create one Trainer (runnable) for each ensemble classifier.
For each arriving data instance (lines 4-17), if the program's elapsed time has not surpassed the set timeout, we obtain the votes from all the classifiers (line 5). 
Then, we compute the \textit{Poisson} weights and update the data structures for training in lines 6-9. The prediction phase has a low computational cost because the algorithm uses Hoeffding trees~\cite{HOEFFDING_TREE}, and thus, we classify instances sequentially. 
On the other hand, the training phase is more expensive. It involves updating many statistics on each tree's nodes, calculating new splits, and detecting data distribution changes (for three methods).
As the training phase dominates the computational cost, parallelism is implemented (in lines 10-13) by simultaneously training many classifiers.
Lines 14-16 represent the global change detector, present only on OBAdwin, where we replace the ensemble's worst classifier with a brand new one.
Finally, lines 17-19 represent the timeout condition, where the ensemble will run for a limited period and then finish processing.

\subsection{Optimizing the performance and power consumption with mini-batching}


Although task parallelism looks straightforward for implementing ensembles, poor memory usage can severely hinder their performance. 
For instance, high-frequency access to data structures larger than cache memories can raise performance bottlenecks. Also, algorithms that continuously perform memory allocation/release operations to discard old models and create new ones during the learning/training process may pressure the garbage collection. 
To mitigate such problems, in a previous work we introduced mini-batching \cite{CASSALES2021260}, an optimization strategy which groups several data instances of a stream for processing.
The algorithm assigns a task to each learner. The tasks' responsibility is to process training by iterating uninterruptedly through all instances of a mini-batch instead of processing a single instance at a time. 
When a task is invoked, its data structures are loaded into the upper levels of the memory hierarchy (upper-level caches). Once on the upper-level caches, the data structures can be quickly accessed to process the remaining instances of the same mini-batch, reducing cache misses and improving performance.

\begin{algorithm}
  \caption{mini-batching algorithm}
  \label{alg:batch}
  \begin{algorithmic}[1]
    \State {\bf Input}: an ensemble $E$, $num\_threads$, a data stream $S$, mini-batch size $L_{mb}$
    \State $P \gets Create\_service\_thread\_pool(num\_threads)$
    \State $T \gets Create\_trainers\_collection(E)$
    \For {each arriving instance $I$ in stream $S$}
    \If {$ElapsedTime > Timeout$}
    \label{alg:iftimeout}
    \State{$E$.process\_minibatch($B$)}
    \State{break loop}
    \EndIf
    \State $B.append(I)$
    \label{alg:mbappend}
    \If{$B.size() == L_{mb}$}
    \label{alg:ifsize}
    \State{$E$.process\_minibatch($B$)}
    \EndIf
    \State sleep();
    \Comment{Sleep until next instance arrives}
    \label{alg:sleep}
    \EndFor
  \end{algorithmic}
\end{algorithm}

The  Algorithm~\ref{alg:batch} shows the mini-batching strategy.
The first difference between the two algorithms appears in lines \ref{alg:iftimeout}-\ref{alg:ifsize} of the Algorithm~\ref{alg:batch}, where the ensemble will only accumulate the instances until the desired mini-batch size is met, the time limit is reached, or the stream ends. If the time limit is reached (line \ref{alg:iftimeout}), the algorithm treats the current mini-batch as if it was complete, processes it, and breaks the execution loop. In normal cases, when the mini-batch size reaches the size set as a parameter (line \ref{alg:ifsize}), the algorithm processes the mini-batch. 
Line \ref{alg:sleep} shows a {\it sleep} operation which is made explicit here just for better illustration on how mini-batching works. In real implementations, such a sleep operation is implicitly implemented by I/O subsystems (e.g., when the algorithm invokes a blocking read operation on a socket to wait for a new incoming data instance). The sleeping period was made explicit here because it is important twofold: ($i$) it releases CPU to other applications (running on the same edge nodes) while waiting for new arriving data instances; ($ii$) it creates opportunity for DPM strategies to turn off idle subsystems (e.g., CPU components) to save energy.  

\begin{algorithm}
  \caption{process\_minibatch routine}
  \label{alg:process}
  \begin{algorithmic}[1]
    \State {\bf Input}: mini-batch $B$
    \For  {each trainer $T_i$ in trainers $T$} {\bf in parallel}
    \label{alg:forclassify}
    \State $T_i.instances \gets B$
    \label{alg:copymb}
    \State $votes_i \gets T_i.classify(T_i.instances)$
    \label{alg:votesclassify}
    \EndFor
    \State $E.compile(votes)$
    \label{alg:compilevotes}
    \For  {each trainer $T_i$ in trainers $T$} {\bf in parallel}
    \label{alg:fortrainparallel}
    \For {each instance $I$ in $T_i.instances$}
    \State $k \gets poisson(\lambda)$
    \State $W\_inst \gets I * k$
    \State $T_i.train\_on\_instance(W\_inst)$
    \EndFor
    \label{alg:endfortrain}
    \If {change detected}
    \label{alg:ifchangedetector}
    \State $reset\_classifier$
    \EndIf
    \label{alg:endifchangedetector}
    \EndFor
    \label{alg:endfortrainparallel}
    \State $B.clear()$
    \label{alg:clear}
  \end{algorithmic}
\end{algorithm}

Algorithm \ref{alg:process} depicts the routine used to process the mini-batch.
It performs the classification (lines \ref{alg:forclassify}-\ref{alg:compilevotes}) and training (lines \ref{alg:fortrainparallel}-\ref{alg:endfortrainparallel}). In line \ref{alg:copymb}, we copy the whole mini-batch to each trainer. Line \ref{alg:votesclassify} uses the instances to compute votes for each trainer and stores them for later use. The votes are aggregated and compiled in line \ref{alg:compilevotes} to provide the predictions. The for in line \ref{alg:forclassify} may be sequential or parallel according to the characteristics of the application (e.g., classifiers with small number of operations may disable the parallelism and run it sequentially). Then, each trainer will iterate (sequentially) through all mini-batch instances while calculating the weight, creating the weighted instance, and training the classifier with this instance (lines \ref{alg:fortrainparallel}-\ref{alg:endfortrain}). ARF, SRP, and LBag, exclusively, will execute lines \ref{alg:ifchangedetector}-\ref{alg:endifchangedetector} as a local change detector for each classifier in the ensemble. 
In OBAdwin, lines \ref{alg:ifchangedetector}-\ref{alg:endifchangedetector} would be outside the parallel section, as the change detection is a global operation.
Finally, in line \ref{alg:clear}, the mini-batch is emptied to begin accumulating again.


In essence, the grouping of instances and reordering of operations improve the algorithm's execution time and energy consumption thanks to a better access locality.
Among the definitions of access locality provided by YUAN et al. \cite{yuan2019relational}, the definition of \textit{reuse distance} (RD) can be used to demonstrate how the mini-batching approach can improve ensemble implementations' access locality. Reuse distance (RD) is defined as ``the number of distinct data accessed since the last access to the same datum, including the reused datum”~\cite{yuan2019relational}.

The RD is $\infty$ for its first access.
The minimum RD is one since it includes the reused datum. The maximum RD, denoted by $m$, is the number of distinct data elements in the problem. In our case, $m$ denotes the number of classifiers in the ensemble (ensemble size).
For example, assuming a set of only three elements $a,b,c$, the RD sequence would be $\infty \infty \infty$ $333$ $333$ for the access sequence \textit{abc abc abc}. As another example, the RD sequence would be $\infty \infty \infty$ $135$ $135$ for the access sequence \textit{abc cba abc}. Therefore,  disregarding the $\infty$ accesses, we can estimate the total RD for the sequential approach as follows:

\begin{equation}
    \label{eq:RDseq}
    RD_{sequential}  =  \sum_{1}^{n} \sum_{1}^{m}m 
\end{equation}

Using the same three-element set from the previous example, the access sequence when using mini-batching would change to \textit{aaa bbb ccc}. In turn, the RD sequence would change to $\infty 1 1$ $\infty 1 1$ $\infty 1 1$ for the first mini-batch and $m 1 1$ for all the subsequent mini-batches. Again, disregarding the $\infty$ accesses, we can estimate the total RD for the mini-batching strategy as follows:

\begin{equation}
    \label{eq:RDmb}
    RD_{mini-batching}  =  \sum_{1}^{\nicefrac{n}{b}} \sum_{1}^{m} (m+b-1)
\end{equation}  

As can be noted by comparing equations \ref{eq:RDseq} and \ref{eq:RDmb}, the most significant contribution of the mini-batching strategy in reducing the RD occurs in the outer sum. By dividing the higher limit by the mini-batch size, a mini-batch size as small as ten can reduce the order of magnitude of this problem. 
At first glance, bigger mini-batches provide a much larger reduction in RD. However, a sufficiently big $n$ (at least, $n > bm^2$) is needed for the outer sum to overcome the impact of the inner sum. 
For more information on this topic, refer to a previous work ~\cite{IS}.

{
In summary, by reducing the RD, mini-batching reduces the number of CPU cycles (and ultimately the energy consumption) to process each data instance. However, notice that mini-batching can improve energy efficiency in a second way. The {sleep} operation in line 4 of Algorithm \ref{alg:sleep} releases the CPU while waiting for the next data instance to arrive. Although this semantics can be implicit in many mini-batching implementations (e.g., our implementation  blocks reading data instances from a socket), it was explicitly included in our algorithm description. These sleep periods put system components in idle state so letting DPM techniques implemented in modern processors to turn off components in order to save energy. Besides saving energy, sleep periods free the processor, which is useful to benefit other applications running on the same physical nodes in the edge.}

\section{Experimental setup}
\label{sec:experiments}

This section describes the experimental evaluation of our proposal. 


\subsection{The testbed}
\label{subsec:environment}

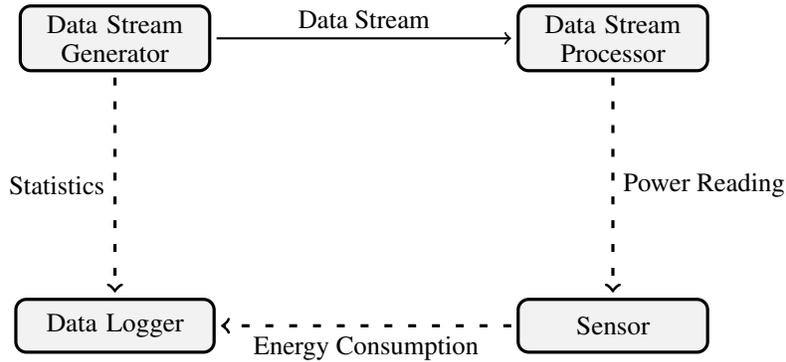
\begin{figure*}[ht]
    \centering
    \begin{tikzpicture}[
    roundrect/.style={
        rectangle,
        rounded corners,
        draw=black, very thick,
        text width=6.5em,
        minimum height=2em,
        text centered,
        fill=black!5},
    data/.style={
        ->,
        thick,
        shorten <=2pt,
        shorten >=2pt,
        },
    monitor/.style={
        dashed,
        very thick, 
        loosely dashed, 
        shorten <=2pt,
        shorten >=2pt,
        ->,
    }
    ]
        \node[roundrect] (mw) {Sensor};
        \node[roundrect, inner sep=5pt,left=4cm of mw] (dlog) {Data Logger};
        \node[roundrect, above=3cm of mw] (proc) {Data Stream Processor};
        \node[roundrect, above=3cm of dlog] (gen) {Data Stream Generator};
        \draw[monitor] (proc) to[out=270,in=90] node[auto]{Power Reading} (mw) ;
        \draw[monitor] (gen) to[out=270,in=90] node[left]{Statistics} (dlog) ;
        \draw[monitor] (mw) to[out=180,in=0] node[auto]{Energy Consumption} (dlog) ;
        \draw[data] (gen) to[out=0,in=180] node[auto]{Data Stream} (proc);
    \end{tikzpicture}
    \caption{The testbed is composed of four nodes (clockwise order): (a) a data stream (load) generator; (b) a data stream processor (whose architectures are described in Table \ref{tab:hard_spec}; (c) a high precision power meter (sensor); and (d) a data logger which registers all experimental data. }
    \label{fig:setup-energy}
\end{figure*}

{
As EC implementations can encompass different hardware platforms ranging from small, low-end devices to mid-end computing servers, we evaluated the optimizations proposed on three hardware architectures. 
The first hardware is a single board computer Raspberry Pi 3 Model B with a Broadcom BCM2837 processor of 4 cores Cortex-A53 64-bit SoC@1.2GHz and 1GB LPDDR2 SDRAM memory, which is frequently pointed as a representative platform for EC implementations. We also included two more powerful hardware platforms, a regular personal computer with Intel i5-2400 CPU@3.10GHz and 4GB memory, and a rack mountable mid-end server SUPERMICRO SYS-7049GP-TRT with dual Xeon CLX-SP 4208 8C/16T, 128GB RAM, dual power supply as described in Table \ref{tab:hard_spec}.  We carried out the experiments on a testbed composed of four nodes (as depicted in  Fig. \ref{fig:setup-energy} connected by a dedicated network.}

\begin{table*}[ht]
\centering
  \caption{ Hardware specifications}
  \label{tab:hard_spec}
  \begin{tabular}{r|ccc}
    Machine type          & single-board & personal & mid-end  \\
                          & computer & computer & server \\
    
    \hline
    Processor        & Intel Xeon 4208 & Intel i5-2400 & Broadcom BCM2835 \\
    Micro architecture   & 	Cascade Lake  & Sandy Bridge & Cortex-A53 \\
    Cores/socket       & 8  & 4 & 4 \\
    Threads/core       & 2  & 1 & 1 \\
    Clock frequency (GHz)  & 2.1 & 3.1 & 1.2 \\
    \hline
    L1 cache (core)    & 32 KB & 128 KB & 32 KB \\
    L2 cache (core)    & 1024 KB & 1024 KB & 512 KB \\
    L3 cache (shared)  & 11264 KB & 6144 KB & - \\
    \hline
    Memory (GB)        & 128  & 4 & 1 \\
    Memory channels    & 6 & 2 & -\\
    Maximum  bandwidth & 107.3 GiB/s & 21 GB/s & - \\
    \hline
    TDP                 &  85 W & 35 W & 4 W \\

\end{tabular}
\end{table*}

\subsection{Load generation}

{
The {\it data stream generator} 
reads the benchmark dataset and transmits the data samples over the network to the {\it data stream processor} node at controlled transmission rates (e.g. to generate low, moderate, and high workloads for each CPU type).} We used five open access \footnote{Available at \url{https://github.com/hmgomes/AdaptiveRandomForest}} datasets  (whose characteristics are summarized in  Table \ref{tab:datasets}) in the experiments:
\begin{table*}[htpb]
    \centering
    \caption{Summary of dataset statistics}
    \label{tab:datasets}
    \begin{tabular}{r|ccccc}
         Datasets & Airlines & GMSC & Electricity & Covertype & Kyoto\\ \hline
         \# Instances & 540k & 150k & 45k & 581k & 725k\\
         \# Features & 7 & 10 & 8 & 54 & 12\\
         \# Nominal feat & 4 & 0 & 1 & 45 & 0\\
         Normalized & No & No & Yes & Yes & Yes\\
    \end{tabular}

\end{table*}

\begin{itemize}
    \item The regression dataset from Ikonomovska inspired the Airlines dataset. The task is to predict whether a given flight will be delayed, given information on the scheduled departure. Thus, it has two possible classes: delayed or not delayed. 
    \item The Electricity dataset was collected from the Australian New South Wales Electricity Market, where prices are not fixed. These prices are affected by the demand and supply of the market itself and set every 5 min. The Electricity dataset tries to identify the price changes (two possible classes: up or down) relative to a moving average of the last 24h. An essential aspect of this dataset is that it exhibits temporal dependencies.
    
    \item The give me some credit (GMSC) dataset is a credit scoring dataset where the objective is to decide whether a loan should be allowed. This decision is crucial for banks since erroneous loans lead to the risk of default and unnecessary expenses on future lawsuits. The dataset contains historical data on borrowers.

    \item The forest covertype dataset represents forest cover type for 30 x 30 m cells obtained from the US Forest Service Region 2 resource information system (RIS) data. Each class corresponds to a different cover type. The numeric attributes are all binary. 
    Moreover, there are seven imbalanced class labels.
    
    \item The Kyoto dataset is an IDS dataset created by researchers from the University of Kyoto. The task is to predict if a flow is an attack of regular traffic. They used honeypots composed of many devices like servers, printers, and IP cameras, among others.
    
\end{itemize}

\subsection{The ensembles used for benchmarking}
\label{sec:testbed}

{
The {\it data stream processor} implements the optimizations as a wrapper for six ensemble algorithms described in section \ref{sec:ensembles}. We implemented this module in the Massive Online Analysis (MOA) framework~\cite{bifet2010moa} \footnote{Avail. at \url{https://github.com/Waikato/moa}}. We adapted MOA to read from a socket instead of reading from a local ARFF file. We chose MOA because: 
($i$) it provides correct and validated implementations of the six ensemble learners used in our experiments,  
($ii$) MOA an be easily extended or modified, which  allowed us to write a wrapper for a uniform evaluation of the six ensembles with the optimizations; and ($iii$) MOA has been used for many studies in the ML area \cite{bifet2010moa}, so that our results can be easily reproduced and compared. The {\it data stream processor} is executed on each different hardware to evaluate its performance and power consumption. }

\subsection{Performance and power consumption measurements}
\label{subsec:evalmeasure}

The {\it data logger} (Fig. \ref{fig:setup-energy}) collects all experimental data  regarding the performance (e.g., throughput, processing delay) and the power consumed by the {data stream processor} for further analysis. The {\it sensor} is implemented by a high precision power meter (Yokogawa MW-100) which periodically collects information directly from the Power Distribution Unit (PDU) and sends to the data logger.

Our interest is to measure the amount of Energy (E) consumed to perform classification tasks. However, most electricity consumption monitors operate by collecting an instantaneous rate of Power (P) being supplied. Energy is the product of the average power and Time (t):
\begin{equation}
    E = P \times t
\end{equation}
Since power can vary in time, the total amount of energy consumed to perform a task is given by
\[ E=\int_{0}^{t} P(t) \,dt \]
where $t$ is the time to perform one task. In practice, we can obtain an approximation of E by taking several periodic measures:
\begin{equation}
    E = \frac{1}{n}\sum_{i=1}^{n}P_i \times  t,
\end{equation}
where $n$ is the number of samples taken by the monitor.
Typically, {\it energy efficiency} is defined as the ratio of the energy spent and the amount of computing performed. As energy is expressed in Joules (J), and the work is expressed as the number of data instances processed (I), we can estimate energy efficiency (as Joules per Instance -  JPI) by:
\begin{equation}
    JPI = \dfrac{E}{I}.
\end{equation}

In some experiments, we use performance metrics of {\it throughput}, given by the average number of data instances processed by second (IPS), and {\it delay}, which is the average time taken to perform the processing of data instances, including its the transmission over the network, the composition of mini-batches, and the time to process the whole mini-batch by the {\it data stream processor}.

The previous metrics are related to computational performance. A second dimension of performance we considered is the {\it predictive performance}, which can be hindered by the optimization techniques. {\it Accuracy} is a widely known measure and represents the percentage of correctly classified instances. 




















{
\section{Experimental results and analysis}
}
\label{subsec:prelim}

\begin{figure*}[t!]
    \centering
    \includegraphics[scale=0.25]{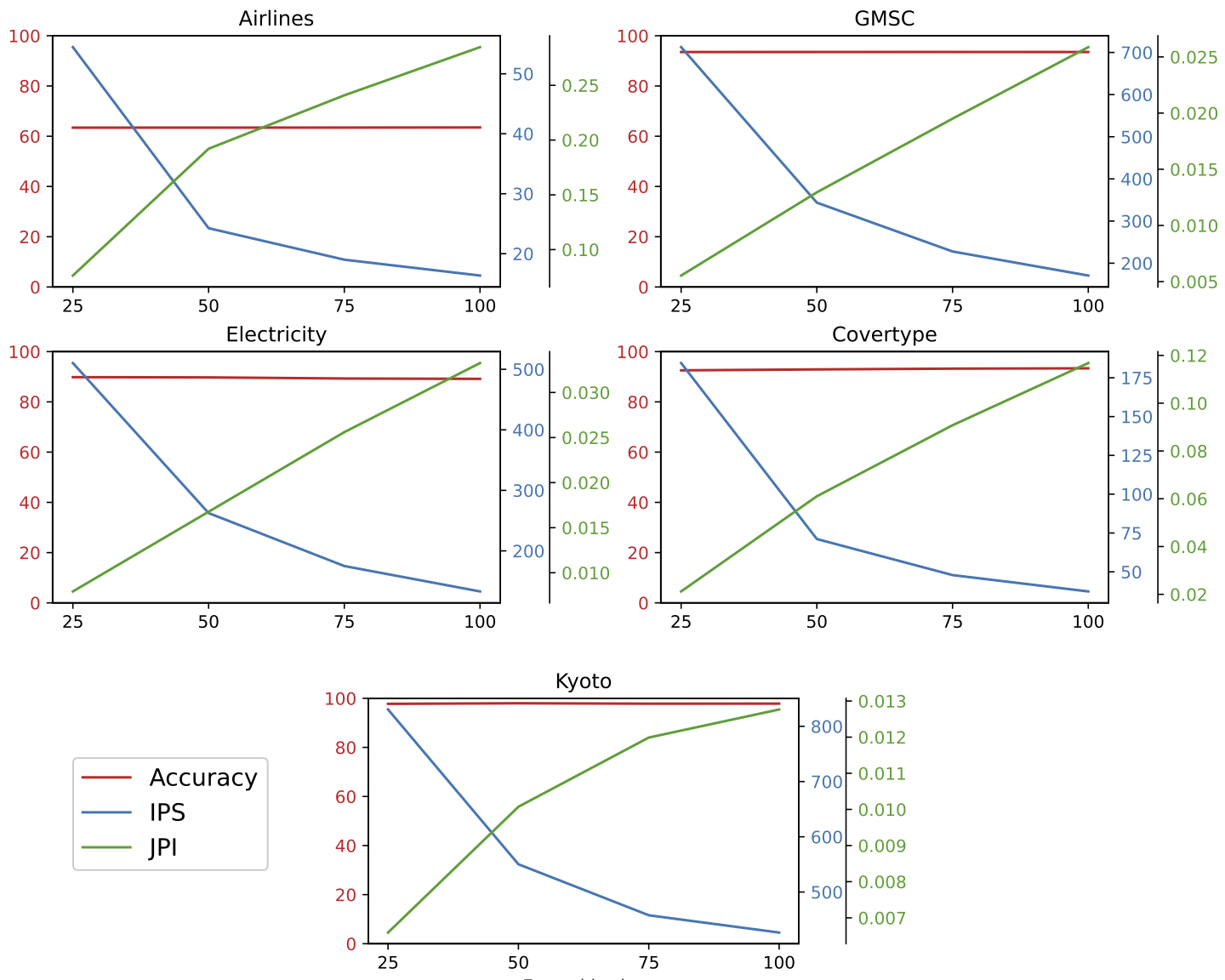}
    \caption{The influence of the ensemble size in accuracy (red), energy efficiency (green), and throughput (blue) for each dataset executing the LBag algorithm on the Raspberry Pi.}
    \label{fig:sizeVS3}
\end{figure*}

As memory constraints of small devices can hinder the execution of large ensembles, our preliminary experiment aimed to find the influence of the ensemble size (i.e., the number of classifiers in the ensemble) on the accuracy, energy consumption, and throughput.  For this experiment, we used the baseline version of the algorithms while measuring energy consumption. Results shown in Fig. \ref{fig:sizeVS3} demonstrate that accuracy (in red) remains almost constant (less than 1\% loss of accuracy), the throughput (in blue) decreases, and energy consumption (in green) increases as we increase the ensemble size (x-axis). Because accuracy loss is negligible for all datasets and all algorithms, used small ensembles in the Raspberry Pi experiments. 

\begin{figure*}[t!]
    \centering
    \begin{subfigure}[t]{0.3\textwidth}
        \centering
        \includegraphics[width=\textwidth]{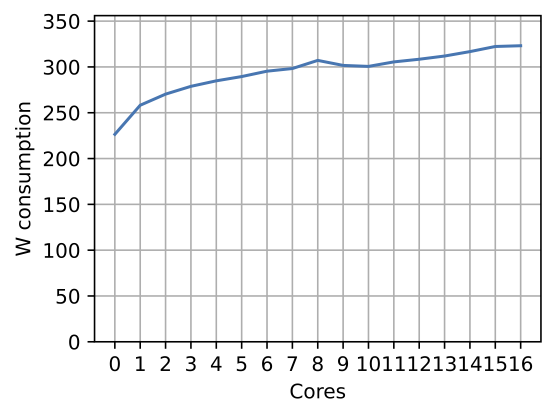}
        \caption{Xeon 4208}
    \end{subfigure}
    ~
    \begin{subfigure}[t]{0.3\textwidth}
        \centering
        \includegraphics[width=\textwidth]{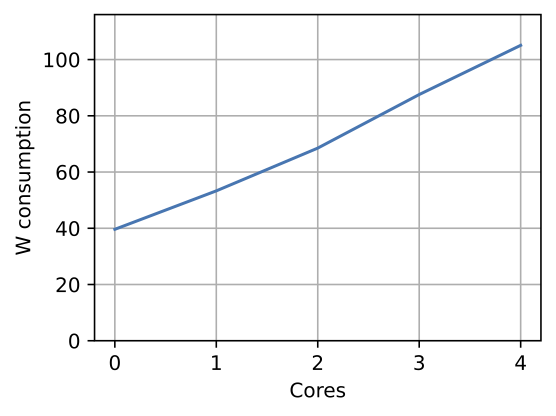}
        \caption{i5-2400}
    \end{subfigure}
    ~
    \begin{subfigure}[t]{0.3\textwidth}
        \centering
        \includegraphics[width=\textwidth]{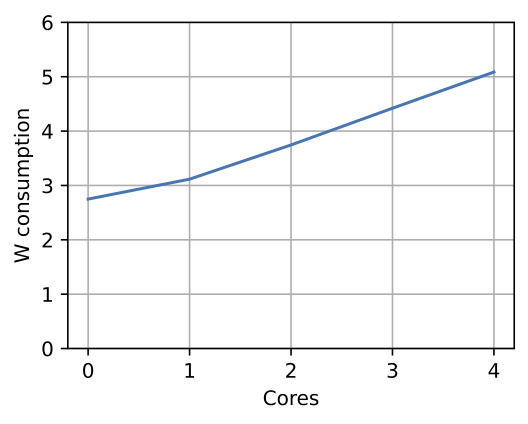}
        \caption{Pi 3 model B}
    \end{subfigure}
    \caption{Energy consumption profile for each machine.}
    \label{fig:energy_profile}
\end{figure*}

The next experiment aims at profiling the power consumption of each machine type as we vary the number of CPU cores running at full load. We used the {\tt stress} application and thread pinning to fully load cores during 180 seconds. 
Results in Fig. \ref{fig:energy_profile} confirm our expectation for the RaspBerry Pi and Core i5-2400. 
Although the TDP for the Xeon 4208 is only 85 W \footnote{\url{https://ark.intel.com/content/www/us/en/ark/products/193390/intel-xeon-silver-4208-processor-11m-cache-2-10-ghz.html}}, the valued measured is higher because it refers to the whole machine (including disk, dual power supply, dual socket, 4 fans, etc.). 

\subsection{Energy consumption}
\label{subsec:sockets}

The next experiment evaluates the energy efficiency  of algorithms running on the three machines under three workload intensities (i.e., at 10\%, 50\%, and 90\% of its maximum throughput). To accomplish that, we used adapted MOA to read data instances from a socket instead of an ARFF file as described in Section \ref{sec:testbed}.
Then, we tuned the load generator to deliver instances at rates equivalent to to 10\%, 50\%, and 90\% of the maximum capacity of the machine for each scenario. Experiments executed each scenario for at least 3 minutes in order to guarantee that the whole system achieved its steady state. 

\begin{figure*}[ht]
    \centering
    \advance\leftskip-1cm
    \includegraphics[width=1.0\linewidth]{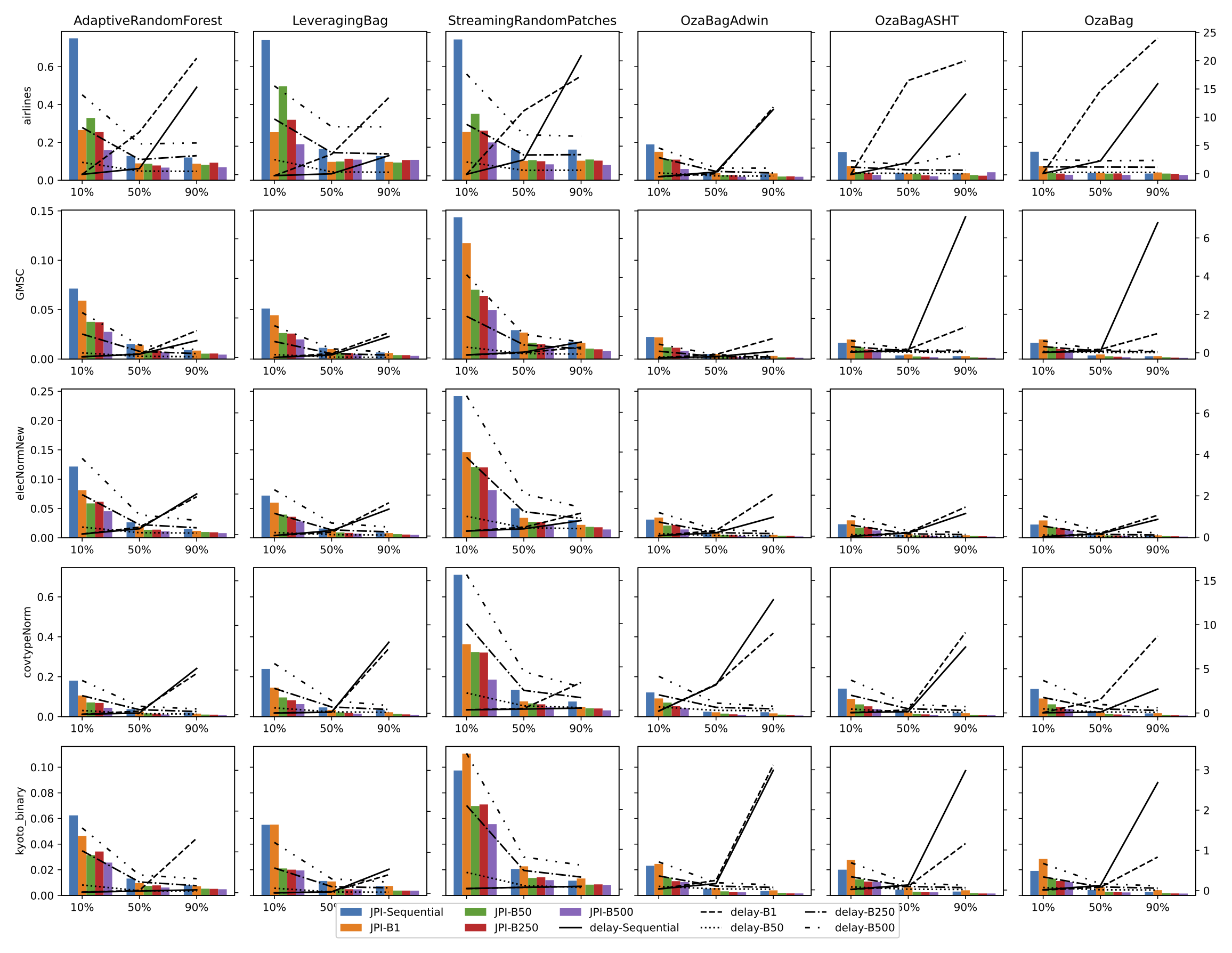}
    \caption{Energy consumption and delay for the Raspberry Pi}
    \label{fig:pi-jpi-delay}
\end{figure*}

\begin{figure*}[ht]
    \centering
    \advance\leftskip-1cm
    \includegraphics[width=1.0\linewidth]{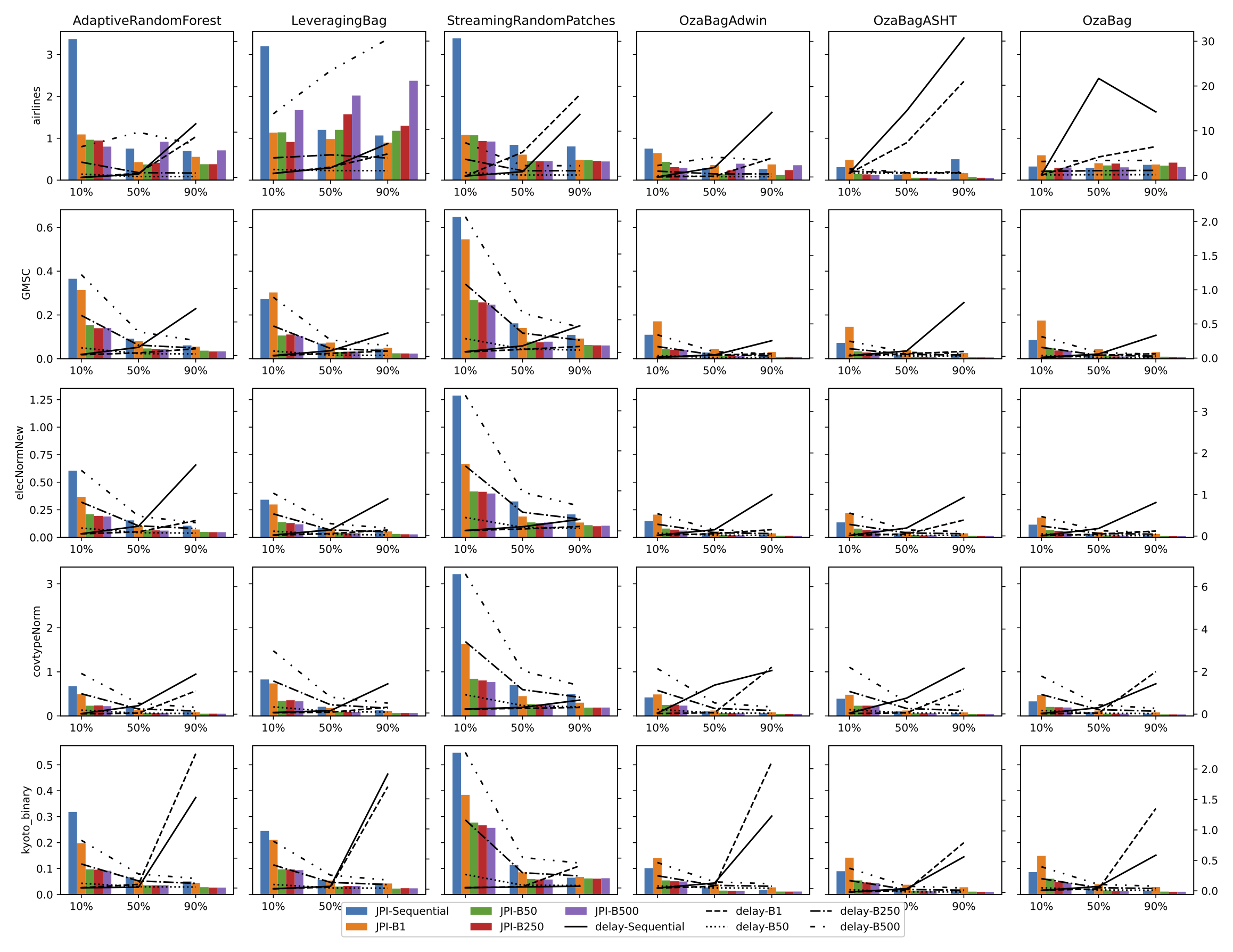}
    \caption{Energy consumption and delay for the i5-2400}
    \label{fig:vostro-jpi-delay}
\end{figure*}

\begin{figure*}[ht]
    \centering
    \advance\leftskip-1cm
    \includegraphics[width=1.0\linewidth]{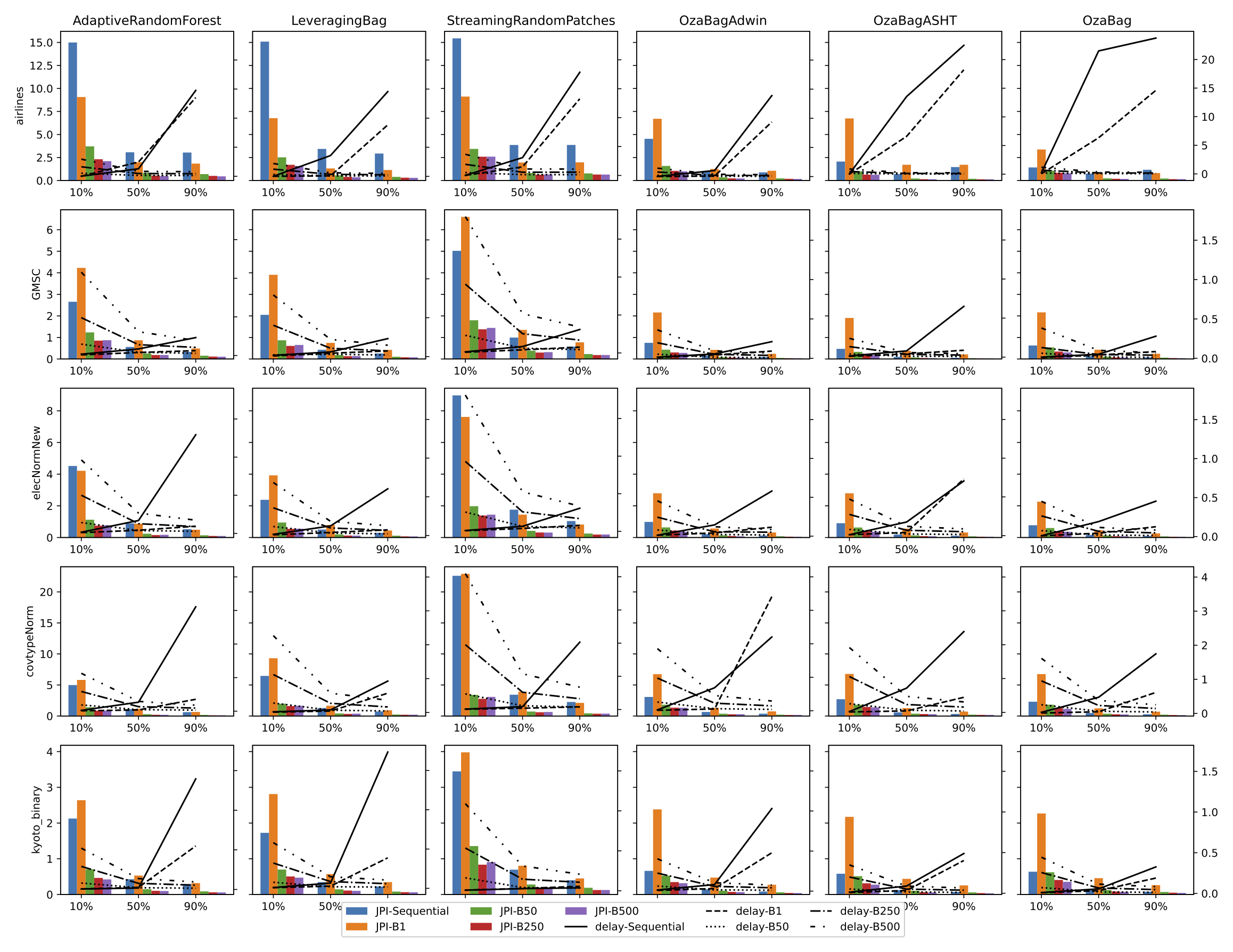}
    \caption{Energy consumption and delay for the Xeon 4208}
    \label{fig:xeon-jpi-delay}
\end{figure*}

In this experiment we compare the baseline (Sequential) implementation of each algorithm, a parallel (multi-thread) implementation without mini-batching (B1), and three parallel versions with mini-batches of 50, 250, and 500 instances (B50, B250, and B500) respectively. Figures \ref{fig:pi-jpi-delay}, \ref{fig:vostro-jpi-delay}, and \ref{fig:xeon-jpi-delay} present the results from Raspberry Pi, i5, and Xeon 4208, respectively. The chart shows results for one dataset per row, whereas the columns show results for each algorithm. All charts in the same row have the same scale. The energy consumption (in Joules per instance - JPI) on the left Y-axis, whereas the average Delay (in milliseconds) per instance appears on the right Y-axis.

As a general remark, energy efficiency of each algorithm varies according to its model complexity. For instance, all three versions of OzaBag remarkably show a better energy efficiency (i.e., smaller JPI) than the other algorithms, because OzaBag produce smaller decision trees that require fewer operations and allow a faster traversal. On the other hand, more complex algorithms presented (proportionally) higher reduction in energy consumption (JPI) when compared to its best counterpart without mini-batching. This behavior is related to the higher throughput produced by the mini-batching version, which shortens the execution time and allows longer periods of low power consumption. Thus, the energy efficiency gains can be explained as follows:
\begin{itemize}
    \item Notice that the mini-batch is processed only in lines 6 and 11 of Algorithm 2, which are executed only when either the time elapsed exceeds the {\it timeout} or the mini-batch is full (i.e., it reaches $L_{mb}$ instances). Otherwise, the loop (line 4) just appends the data instance 
    in the mini-batch and waits (i.e., it enters in a sleep state) for the arrival of the next data instance. Because the thread sleeps for a while, it does not consume CPU cycles, so letting the DPM mechanisms implemented by the CPU hardware to turn off idle system components and thus save power;
    \item Even under high loads (i.e., sleep periods may tend to zero), mini-batching yields power reduction because it reduces cache misses (as demonstrated in \cite{CASSALES2021260}), thus accelerating the stream processing and reducing the number of cycles and memory accesses per instance.
\end{itemize}

Although mini-batching can improve energy efficiency, the delay resulting from extended and repeated sleep periods may hinder the idea of real-time processing, primarily when the rate of incoming instances is several times smaller than the mini-batch size.
Such a phenomenon appears only in the B250 and B500 experiments.
On the other hand, the B50 experiments present delays similar to the instance-by-instance implementations (Seq and B1) while having a better energy efficiency.

Regardless of the platform and dataset used, SRP has the worst energy efficiency across all the experiments. It happens because this algorithm produces higher decision trees than the other complex algorithms (e.g., ARF and LBag), thus increasing the computational complexity for the stream processing.

Remarkably, mini-batches of 50 near instances yield better benefits in terms of delay reduction, energy efficiency, and smaller impact in predictive performance, while larger mini-batches than this size presented diminishing returns. This result is consistent to our previous work in \cite{CASSALES2021260}.


More detailed results on the energy efficiency gains are given in Tables \ref{tab:delta-Pi}, \ref{tab:delta-Vostro}, and 
\ref{tab:delta-Xeon}. They compare energy consumption between the best version without mini-batch (i.e., the best case between the orange and blue bars in the charts) and the mini-batch version. Negative values indicates the (percentage) reduction in energy consumption due to mini-batching. Cases where mini-batch consumed more energy have positive values and are in bold. 

\begin{table*}[ht]
\centering
\advance\leftskip-1cm
\setlength\tabcolsep{2.1pt}
\begin{tabular}{c|ccc|ccc|ccc|ccc|ccc}
& \multicolumn{3}{c|}{Airlines} & \multicolumn{3}{c|}{GMSC} & \multicolumn{3}{c|}{Electricity} & \multicolumn{3}{c|}{Covertype} & \multicolumn{3}{c}{Kyoto}\\
Algorithm &  10       & 50      & 90      & 10     & 50     & 90     & 10        & 50       & 90       & 10       & 50       & 90   & 10       & 50       & 90    \\
\hline

ARF & -39.86 & -24.87 & -21.43 & -53.42 & -51.29 & -47.81 & -44.05 & -42.69 & -32.69 & -58.19 & -56.56 & -52.42 & -44.83 & -35.61 & -32.79  \\
LBag & -25.04 & \textbf{ 11.88} & \textbf{ 10.89} & -55.20 & -51.70 & -47.94 & -53.48 & -45.95 & -39.18 & -56.66 & -55.94 & -57.63 & -64.64 & -56.52 & -48.65  \\
SRP & -21.66 & -18.04 & -22.09 & -57.89 & -55.78 & -51.54 & -44.29 & -37.97 & -35.43 & -48.95 & -38.37 & -36.49 & -42.91 & -42.20 & -28.70  \\
OBAd & -60.30 & -51.74 & -48.96 & -61.54 & -57.91 & -52.18 & -53.96 & -50.63 & -44.30 & -55.02 & -56.95 & -64.73 & -54.97 & -53.28 & -54.20  \\
OBASHT & -62.30 & -40.61 & \textbf{ 17.97} & -55.46 & -50.67 & -58.96 & -45.85 & -40.91 & -33.80 & -56.95 & -56.22 & -60.48 & -47.45 & -45.95 & -50.04  \\
OB & -61.23 & -27.51 & -21.63 & -55.14 & -51.26 & -58.31 & -44.83 & -39.42 & -34.47 & -57.53 & -56.68 & -65.04 & -45.84 & -44.35 & -46.95  \\

\end{tabular}
\caption{Percentage difference between the best non-mini-batch version and the mini-batch version on the Raspberry Pi}
\label{tab:delta-Pi}
\end{table*}

\begin{table*}[ht]
\centering
\advance\leftskip-1cm
\setlength\tabcolsep{2pt}
\begin{tabular}{c|ccc|ccc|ccc|ccc|ccc}
& \multicolumn{3}{c|}{Airlines} & \multicolumn{3}{c|}{GMSC} & \multicolumn{3}{c|}{Electricity} & \multicolumn{3}{c|}{Covertype} & \multicolumn{3}{c}{Kyoto}\\
Algorithm &  10       & 50      & 90      & 10     & 50     & 90     & 10        & 50       & 90       & 10       & 50       & 90   & 10       & 50       & 90    \\
\hline
ARF & -26.64 & \textbf{ 112.66} & \textbf{ 27.88} & -55.09 & -45.61 & -39.58 & -48.65 & -40.06 & -34.12 & -55.05 & -48.11 & -42.40 & -53.61 & -34.79 & -36.08  \\
LBag & \textbf{ 47.50} & \textbf{ 105.85} & \textbf{ 164.66} & -62.05 & -52.71 & -48.87 & -60.39 & -50.86 & -49.39 & -54.99 & -48.19 & -43.84 & -55.43 & -38.61 & -43.88  \\
SRP & -14.89 & -25.45 & -8.34 & -54.75 & -44.62 & -34.29 & -40.39 & -28.94 & -21.54 & -52.98 & -42.79 & -36.58 & -33.10 & -32.57 & -3.14  \\
OBAd & -54.64 & \textbf{ 36.33} & \textbf{ 34.65} & -63.66 & -57.72 & -55.74 & -58.95 & -54.26 & -52.44 & -44.37 & -39.91 & -33.65 & -50.78 & -47.40 & -37.44  \\
OBASHT & -60.56 & -58.60 & -67.16 & -63.21 & -59.05 & -51.85 & -54.07 & -47.84 & -45.11 & -40.51 & -39.83 & -35.71 & -50.43 & -46.33 & -36.05  \\
OB & -7.19 & \textbf{ 5.56} & -13.34 & -57.65 & -54.94 & -46.74 & -51.92 & -48.58 & -45.28 & -42.21 & -40.24 & -33.24 & -47.55 & -42.07 & -30.06  \\
\end{tabular}

\caption{Percentage difference between the best non-mini-batch version and the mini-batch version on the i5-2400}
\label{tab:delta-Vostro}
\end{table*}

\begin{table*}[ht]
\centering
\advance\leftskip-1cm
\setlength\tabcolsep{2.5pt}
\begin{tabular}{c|ccc|ccc|ccc|ccc|ccc}
& \multicolumn{3}{c|}{Airlines} & \multicolumn{3}{c|}{GMSC} & \multicolumn{3}{c|}{Electricity} & \multicolumn{3}{c|}{Covertype} & \multicolumn{3}{c}{Kyoto}\\
Algorithm &  10       & 50      & 90      & 10     & 50     & 90     & 10        & 50       & 90       & 10       & 50       & 90   & 10       & 50       & 90    \\
\hline
ARF & -76.81 & -75.08 & -75.30 & -67.09 & -66.98 & -67.38 & -81.74 & -80.47 & -80.30 & -82.64 & -83.06 & -82.33 & -80.12 & -77.36 & -80.25  \\
LBag & -76.71 & -73.22 & -74.66 & -68.16 & -68.30 & -68.92 & -76.63 & -76.37 & -77.18 & -74.41 & -71.04 & -71.24 & -72.75 & -70.85 & -72.61  \\
SRP & -71.57 & -66.02 & -67.11 & -71.23 & -67.91 & -67.84 & -81.06 & -78.40 & -76.87 & -86.37 & -81.20 & -82.01 & -73.70 & -71.24 & -68.50  \\
OBAd & -78.07 & -77.87 & -80.59 & -62.25 & -63.51 & -63.70 & -63.50 & -64.24 & -64.56 & -56.87 & -55.37 & -55.34 & -52.19 & -54.16 & -51.48  \\
OBASHT & -69.59 & -79.84 & -90.91 & -60.48 & -61.67 & -62.04 & -59.17 & -58.30 & -61.89 & -48.61 & -49.38 & -48.26 & -52.69 & -53.35 & -52.64  \\
OB & -44.22 & -77.02 & -81.20 & -51.25 & -52.18 & -52.74 & -53.49 & -54.06 & -57.95 & -50.19 & -48.38 & -49.19 & -44.46 & -45.90 & -43.83  \\
\end{tabular}
\caption{Percentage difference between the best non-mini-batch version and the mini-batch version on the Xeon}
\label{tab:delta-Xeon}
\end{table*}

\begin{figure*}[ht]
    \centering
    \advance\leftskip-1cm
    \includegraphics[width=1.0\linewidth]{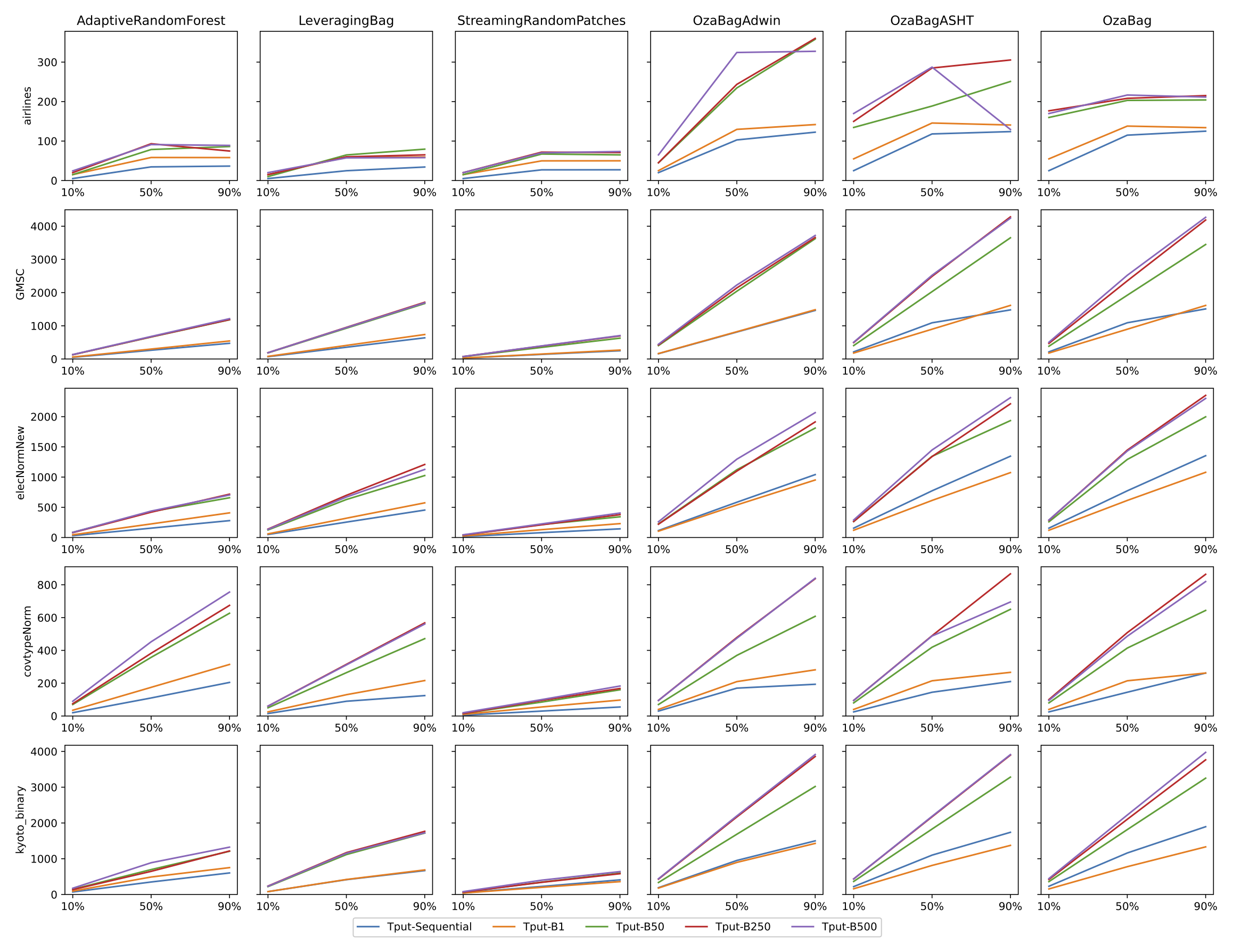}
    \caption{Throughput for the Raspberry Pi}
    \label{fig:pi-tput}
\end{figure*}

\begin{figure*}[ht]
    \centering
    \advance\leftskip-1cm
    \includegraphics[width=1.0\linewidth]{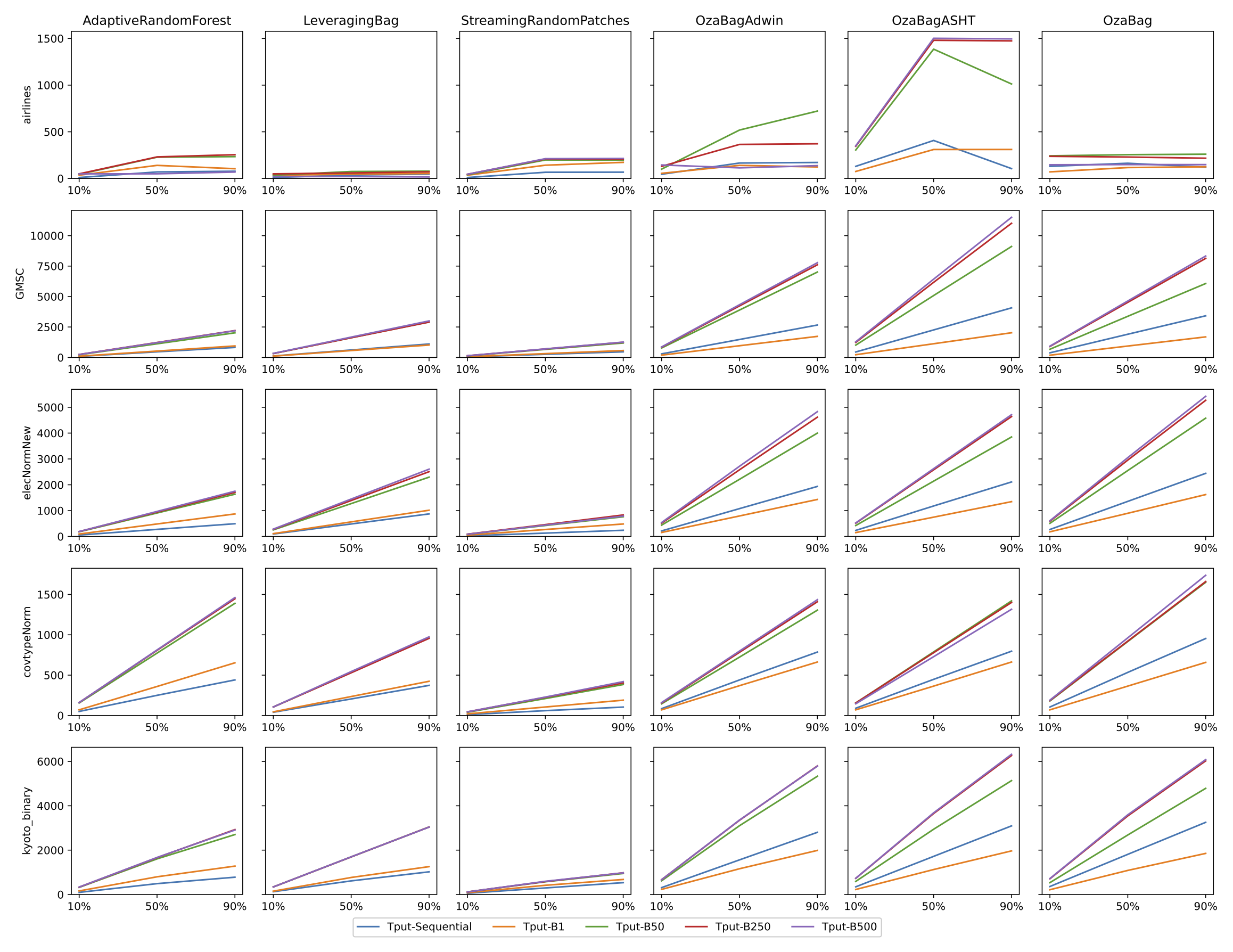}
    \caption{Throughput for the i5-2400}
    \label{fig:vostro-tput}
\end{figure*}

\begin{figure*}[ht]
    \centering
    \advance\leftskip-1cm
    \includegraphics[width=1.0\linewidth]{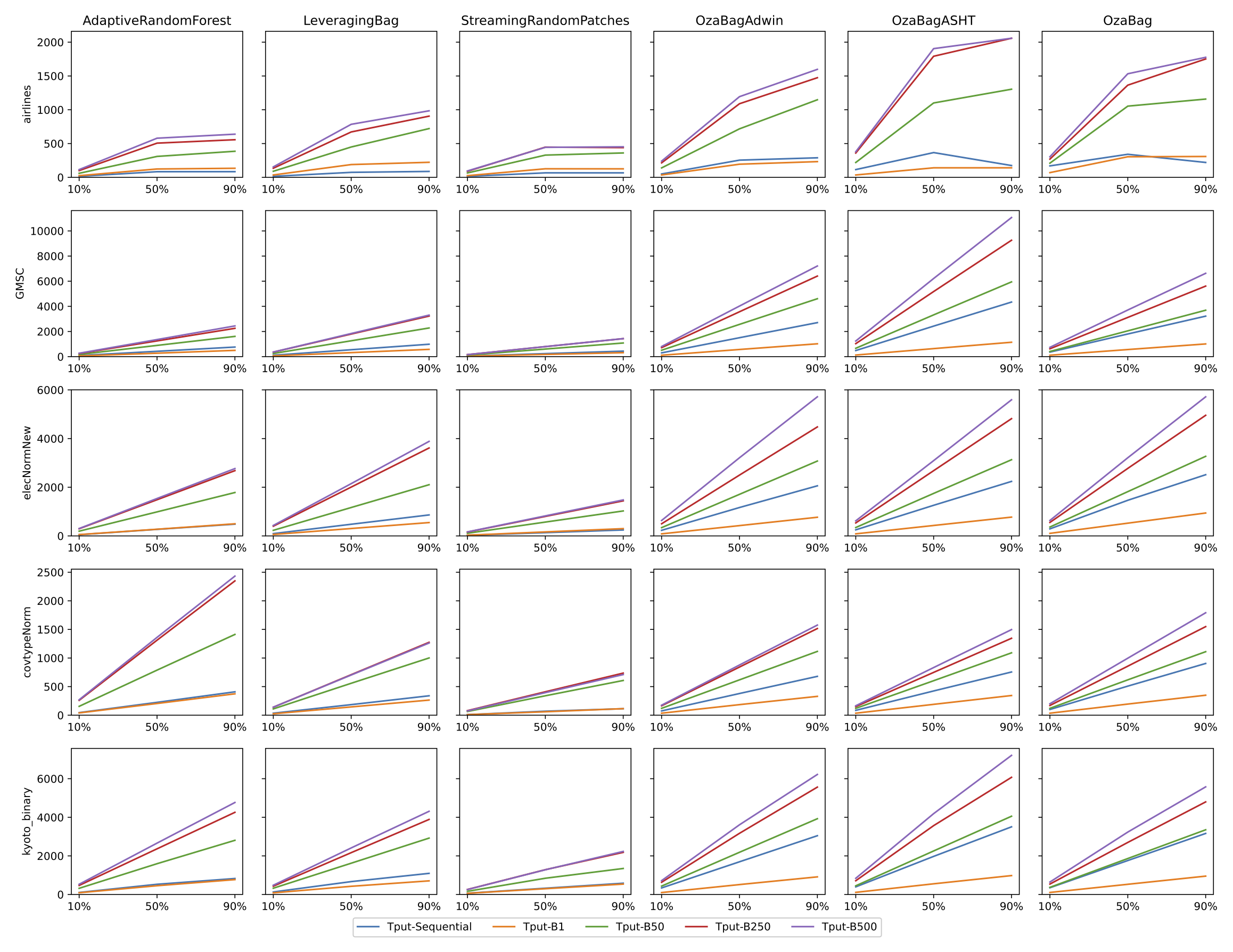}
    \caption{Throughput for the Xeon 4208}
    \label{fig:xeon-tput}
\end{figure*}

It is possible to see in the tables that using mini-batch of 50 instances (MB50) improves both performance and energy efficiency in 259 out of 270 experiments (i.e., 96\%). However, mini-batching has proven to increase the performance (e.g., throughput) in all the 270 experiments as illustrated in Figures \ref{fig:pi-tput}, \ref{fig:vostro-tput}, and \ref{fig:xeon-tput} and in Table \ref{tab:tput-vostro-lbag-airlines}. 



\begin{table}[ht]
\centering
\begin{tabular}{r|c|c|c}
Algorithm & 10\%   & 50\%     & 90\%     \\
\hline
Sequential & 9.99 &	37.05 &	46.56\\
B1 & 34.96 & 45.68 & 53.46 \\
B50 & 34.92 & 74.44 & 78.12 \\
B250 & 49.19 & 58.95 & 71.05 \\
B500 & 20.01 & 20.80 & 16.21\\
\end{tabular}
\caption{Throughput for the algorithm LeveragingBag using the Airlines dataset.}
\label{tab:tput-vostro-lbag-airlines}
\end{table}

In summary, mini-batching can support consistent improvements in energy consumption and time performance across the experiments. First, mini-batching improves the RD, which reduces the number of CPU cycles to process each data instance faster, and spending less energy. Second, the sleeping periods created (while awaiting for new data instances to compose a new mini-batch) create opportunity for DPM strategies to turn off idle subsystem (e.g., many processor subsystems) to save energy. 

Also, notice that mini-batching causes a trade-off between energy consumption, time metrics (delay, and throughput), and accuracy.
While the mini-batching increases the throughput, delay, and energy efficiency, larger mini-batches can modify the accuracy of the algorithms. 
This result corroborated by previous studies on mini-batching performance \cite{IS}. However, as demonstrated in our experiments, it is possible to balance this trade off. The mini-batch size of 50 instances is a good balance demonstrated for practically all the experimental scenarios studied.

\section{Conclusion}
\label{sec:conclusions}

Ensemble learning is a fruitful approach to improve the performance of ML models by combining several single models.
Ensembles are also popular in a data stream processing context, where they achieve remarkable predictive performance. 
Examples of this class include algorithms such as Adaptive Random Forest, Leveraging Bag, and OzaBag. 
Despite their relevance, many aspects of their efficient implementation remained to be studied after their original proposals. 
For example, the original Adaptive Random Forest implementation included a simple multi-thread version, but it did not consider energy efficiency or mini-batches to improve the overall run time. 

In this paper, we proposed an experimental framework to evaluate the performance and energy efficiency of the mini-batching strategy proposed in \cite{IS} under a realistic scenario where data streams are sent through the network at different load intensities, with six
state-of-art ensemble algorithms (OzaBag, OzaBag Adaptive Size Hoeffding Tree, Online Bagging ADWIN,
Leveraging Bagging, Adaptive RandomForest, and Streaming Random Patches) processing five widely used
machine learning benchmark datasets with varied characteristics on three computer platforms. 

Our study demonstrated that 
mini-batching yields remarkable reduction in energy consumption and time performance, at cost of small changes in the predictive performance. Larger improvements in time performance and energy savings were observed at low loads, with a trade-off on the average delay to process the instances. Conversely, mini-batching increases throughput when the workload is high, where the energy-reduction is present (but not as intense) and the delay is smaller than the baseline.
Despite of the trade-offs observed, it is possible to balance them to avoid significant loss in predictive performance.

In future work, we intend to investigate if it is possible to improve the solution by using an adaptive mini-batching size regarding predictive performance, throughput, delay, and energy consumption.

\section{Acknowledgements}
\noindent This study was financed in part by the Coordenação de Aperfeiçoamento de Pessoal de Nível Superior - Brasil (CAPES) - Finance Code 001, and Programa Institucional de Internacionalização – CAPES-PrInt UFSCar (Contract 88887.373234/2019-00). Authors also thank Stic AMSUD (project 20-STIC-09), and FAPESP (contract numbers  2018/22979-2, and 2015/24461-2) for their support. Partially supported by the TAIAO project CONT-64517-SSIFDS-UOW (Time-Evolving Data Science / Artificial Intelligence for Advanced Open Environmental Science) funded by the New Zealand Ministry of Business, Innovation, and Employment (MBIE). URL https://taiao.ai/.

\bibliographystyle{IEEEtran}
\bibliography{sample-base}

\end{document}